\pdfoutput=1

\documentclass[11pt]{article}

\newif\ifreviewmode
\reviewmodefalse %

\ifreviewmode
\usepackage[review]{ACL2023}
\else
\usepackage{ACL2023}
\fi

\usepackage{times}
\usepackage{latexsym}

\usepackage[T1]{fontenc}

\usepackage[utf8]{inputenc}

\usepackage{microtype}

\usepackage{inconsolata}

\usepackage{algpseudocode}
\usepackage{algorithm}

\usepackage{amsmath}
\usepackage{ dsfont }
\usepackage{forest}
\usepackage{amsthm}
\theoremstyle{definition} %

\newtheorem{prop}{Proposition}

\usepackage{cancel}

\newcommand{\expec}[1]{\ensuremath{\mathds{E}_{#1}}}

\usepackage[nocomma]{optidef}

\usepackage[capitalise]{cleveref}

\usepackage{microtype}

\usepackage{xcolor}

\DeclareMathOperator*{\argmax}{arg\:max}
\DeclareMathOperator*{\argmin}{arg\:min}

\usepackage{stmaryrd}

\newcommand{\Perm}{\ensuremath{\mathcal{P}}}

\usepackage{algpseudocode}
\usepackage{algorithm}

\usepackage{xcolor}
\usepackage{xspace}

\usepackage{ amssymb }

\usepackage{booktabs}

\newcommand{\grammar}{\ensuremath{\dagger}\xspace}
\newcommand{\nogrammar}{}

\newcommand{\ie}{i.e.\ }
\newcommand{\eg}{e.g.\ }

\newcommand{\ci}[1]{\scalebox{0.6}{\ensuremath{\textcolor{darkgray}{\pm #1}}}}
\newcommand{\cinf}[1]{\ci{#1}}

\usepackage{ bbold }

\usepackage{ mathrsfs }

\usepackage{multirow}

\newcommand{\pretrained}{\ensuremath{^{\diamond}\xspace}}

\pgfdeclarelayer{background}
\pgfdeclarelayer{main}
\pgfsetlayers{background,main}

\newcommand{\treemodel}{\ensuremath{^{\clubsuit}\xspace}}

\newcommand{\tightmath}{\abovedisplayskip=5pt
\belowdisplayskip=5pt}

\renewcommand{\tightmath}{}

\newcommand{\shortfert}{L'23\xspace}

\usepackage{enumitem}

\setlist[itemize]{parsep=0pt,topsep=2pt,itemsep=0ex,partopsep=1ex,parsep=1ex}

\renewcommand{\paragraph}[1]{\textbf{#1.} }

\newcommand{\dbar}{\ensuremath{\ || \ }}

\newcommand{\kl}{\text{KL}}

\usepackage{accents}

\newcommand{\sscore}[2]{s_{#1 \curvearrowright #2}}

\usepackage{tabularx}

\title{Compositional Generalization without Trees \\ using Multiset Tagging and Latent Permutations}

\author{Matthias Lindemann$^1$ \and Alexander Koller$^2$ \and Ivan Titov$^{1,3}$ \\
$^1$ ILCC, University of Edinburgh,
$^2$ LST, Saarland University,
$^3$ ILLC, University of Amsterdam \\
{\small \texttt{m.m.lindemann@sms.ed.ac.uk}, \texttt{koller@coli.uni-saarland.de}, \texttt{ititov@inf.ed.ac.uk} }
}

\begin{document}
\maketitle

\begin{abstract}

Seq2seq models have been shown to struggle with compositional generalization in semantic parsing, \ie generalizing to unseen compositions of phenomena that the model handles correctly in isolation.
We phrase semantic parsing as a two-step process: we first tag each input token with a multiset of output tokens. Then we arrange the tokens into an output sequence using a new way of parameterizing and predicting permutations. We formulate predicting a permutation as solving a regularized linear program and we backpropagate through the solver. In contrast to prior work, our approach does not place a priori restrictions on possible permutations, making it very expressive.
Our model outperforms pretrained seq2seq models and prior work on realistic semantic parsing tasks that require generalization to longer examples. We also outperform non-tree-based models on structural generalization on the COGS benchmark.
For the first time, we show that a model without an inductive bias provided by trees achieves high accuracy on generalization to deeper recursion depth.\footnote{
\ifreviewmode
(github link redacted)
\else
\url{https://github.com/namednil/multiset-perm}
\fi
}

\end{abstract}

\section{Introduction}

\begin{figure}[t]
    \centering
    \includegraphics[width=\linewidth]{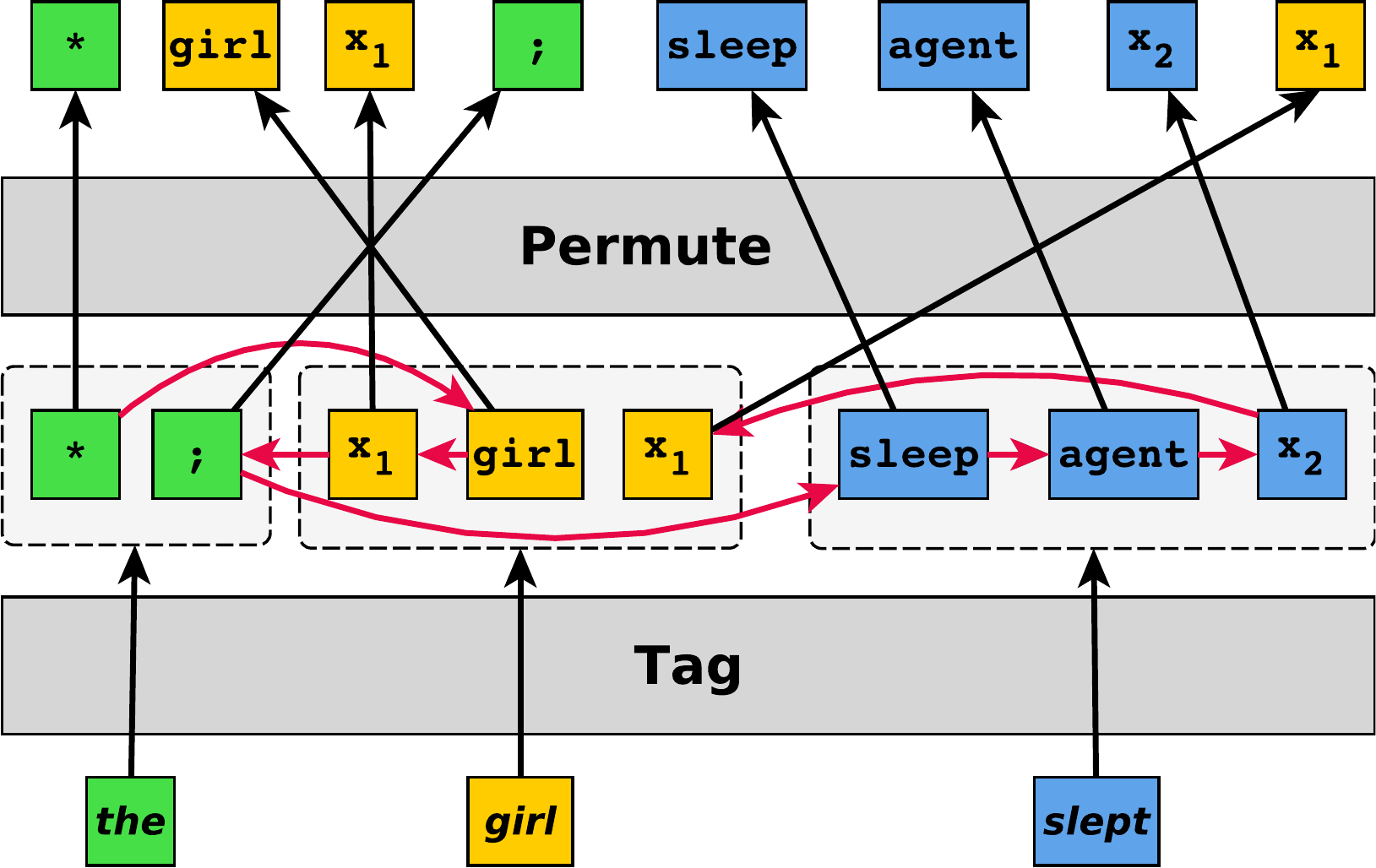}
    \caption{We model seq2seq tasks by first predicting a multiset (dashed boxes) for every input token, and then permuting the tokens to put them into order. We realize the permutation using the red edges: if there is an edge from token $i$ to $j$, then $i$ is the predecessor of $j$ in the output. Every token is visited exactly once.}
    \label{fig:intro}
\end{figure}

Sequence-to-sequence models have been very successfully applied to many structural tasks in NLP such as semantic parsing. However, they have also been shown to struggle with compositional generalization \citep{lake2018generalization, finegan-dollak-etal-2018-improving, kim-linzen-2020-cogs, hupkes2020compositionality}, \ie the model fails on examples that contain unseen compositions or deeper recursion of phenomena that it handles correctly in isolation. For example, a model which correctly parses `Mary knew that Jim slept' should also be able to parse sentences with deeper recursion than it has seen during training such as `Paul said that Mary knew that Jim slept'. This sort of generalization is easy for humans but challenging for neural models.

In order for a model to generalize compositionally in semantic parsing, it has to identify reusable `fragments' and be able to systematically combine them in novel ways. One way to make a model sensitive to fragments is to make it rely on a tree that makes the compositional structure explicit. However, this complicates the training because these trees need to be obtained or induced. This is computationally demanding or at least requires structural preprocessing informed by domain knowledge.

In this paper, we propose a two-step sequence-based approach with a structural inductive bias that does not rely on trees: the `fragments' are multisets of output tokens that we predict for every input token in a first step. 
A second step then arranges the tokens we predicted in the previous step into a single sequence using a permutation model. 
In contrast to prior work \citep{wang2021structured, fertility-paper}, there are no hard constraints on the permutations that our model can predict. We show that this enables structural generalization on tasks that go beyond the class of synchronous context-free languages.

We overcome two key technical challenges in this work: Firstly, we do not have supervision for the correspondence between input tokens and output tokens. Therefore, we induce the correspondence during training. Secondly, predicting permutations without restrictions is computationally challenging. 
For this, we develop a differentiable GPU-friendly algorithm.

On realistic semantic parsing tasks our approach outperforms previous work on generalization to longer examples than seen at training.
We also outperform all other non-tree models on the structural generalization tasks in semantic parsing on COGS \citep{kim-linzen-2020-cogs}. For the first time, we also show that a model without an inductive bias towards trees can obtain high accuracy on generalization to deeper recursion on COGS.

To summarize, our main contributions are:
\begin{itemize}
    \item a flexible seq2seq model that performs well on structural generalization in semantic parsing without assuming that input and output are related to each other via a tree structure.
    \item a differentiable algorithm to parameterize and predict permutations without a priori restrictions on what permutations are possible.
\end{itemize}

\section{Overview and Motivation}
\label{sec:setup}

Our approach consists of two stages.  In the \textbf{first stage} (multiset tagging), we predict a multiset $\textbf{z}_i$ of tokens for every input token $\textbf{x}_i$ from the contextualized representation of $\textbf{x}_i$. This is motivated by the observation that input tokens often systematically correspond to a fragment of the output (like \textit{slept} corresponding to \texttt{sleep} and \texttt{agent} and a variable in \cref{fig:intro}). Importantly, we expect this systematic relationship to be largely invariant to phrases being used in new contexts or deeper recursion. 
We refer to the elements of the multisets as \textit{multiset tokens}. %

In the \textbf{second stage} (permutation), we order the multiset tokens to arrive at a sequential output. Conceptually, we do this by going from left to right over the \textit{output} and determining which multiset token to put in every position. Consider the example in \cref{fig:intro}. For the first output position, we simply select a multiset token (\texttt{*} in the example). All subsequent tokens are put into position by `jumping' from the token that was last placed into the output to a new multiset token. In \cref{fig:intro}, we jump from \texttt{*} to \texttt{girl} (shown by the outgoing red edge from \texttt{*}). This indicates that \texttt{girl} is the successor of \texttt{*} in the output and hence the \textit{second} output token. From \texttt{girl} we jump to one of the $\texttt{x}_{\texttt{1}}$ tokens to determine what the \textit{third} output token is and so on. Since we predict a permutation, we must visit every multiset token exactly once in this process.

The jumps are inspired by reordering in phrase-based machine translation \citep{koehn-etal-2007-moses} and methods from dependency parsing, where directly modeling bi-lexical relationships on hidden states has proven successful \citep{kiperwasser-goldberg-2016-simple}. Note also that \textit{any} permutation can be represented with jumps. In contrast, prior work \citep{wang2021structured, fertility-paper} has put strong restrictions on the possible permutations that can be predicted. Our approach is more flexible and empirically it also scales better to longer inputs, which opens up new applications and datasets. %

\paragraph{Setup} We assume we are given a dataset $\mathcal{D} = \{(\textbf{x}^1, \textbf{y}^1), \ldots\}$ of input utterances $\textbf{x}$ and target sequences $\textbf{y}$. If we had gold alignments, it would be straightforward to train our model. Since we do not have this supervision, we have to discover during training which tokens of the output $\textbf{y}$ belong into which multiset $\textbf{z}_i$.
We describe the model and the training objective of the multiset tagging in \cref{sec:multisets}. After the model is trained, we can annotate our training set with the most likely $\textbf{z}$, and then train the permutation model.

For predicting a permutation, we associate a score with each possible jump and search for the highest-scoring sequence of jumps. We ensure that the jumps correspond to a permutation by means of constraints, which results in a combinatorial optimization problem.
The flexibility of our model and its parametrization come with the challenge that this problem is NP-hard. We approximate it with a regularized linear program which also ensures differentiability. Our permutation model and its training are described in \cref{sec:relaxed-perms}. In \cref{sec:bregman-for-perm}, we discuss how to solve the regularized linear program and how to backpropagate through the solution.

\section{Learning Multisets}
\label{sec:multisets}
For the multiset tagging, we need to train a model that predicts the multisets $\textbf{z}_1, \ldots, \textbf{z}_n$ of tokens by conditioning on the input. We represent a multiset $\textbf{z}_i$ as an integer-valued vector that contains for every vocabulary item $v$ the multiplicity of $v$ in $\textbf{z}_i$, \ie $\textbf{z}_{i,v} = k$ means that input token $i$ contributes $k$ occurrences of output type $v$. If $v$ is not present in multiset $\textbf{z}_i$, then $\textbf{z}_{i,v}=0$. For example, in \cref{fig:intro}, $\textbf{z}_{3, \texttt{sleep}} =1 $ and $\textbf{z}_{2, \texttt{x}_1} = 2$.
As discussed in \cref{sec:setup}, we do not have supervision for $\textbf{z}_1, \ldots, \textbf{z}_n$ and treat them as latent variables. To allow for efficient exact training, we assume that all $\textbf{z}_{i,v}$ are independent of each other conditioned on the input:
\tightmath
\begin{align}
    P(\textbf{z}\mid \textbf{x}) = \prod_{i,v} P(\textbf{z}_{i,v}\mid \textbf{x}) \label{eq:multiset}
\end{align}
where $v$ ranges over the \textit{entire} vocabulary.

\paragraph{Parametrization} We parameterize $P(\textbf{z}_{i,v}\mid \textbf{x})$ as follows. We first pass the input $\textbf{x}$ through a pretrained RoBERTa encoder \citep{Liu2019RoBERTaAR}, where $\textsc{encoder}(\textbf{x})$ is the output of the final layer. We then add RoBERTa's static word embeddings from the first, non-contextualized, layer to that:
\begin{align}
    \textbf{h}_i = \textsc{encoder}(\textbf{x})_i + \textsc{embed}(\textbf{x}_i) \label{eq:encode}
\end{align}
We then pass $\textbf{h}_i$ through a feedforward network obtaining $\tilde{\textbf{h}}_i = \textsc{ff}(\textbf{h}_i)$ and define a distribution over the multiplicity of $v$ in the multiset $\textbf{z}_i$: 
\begin{align}
    P(\textbf{z}_{i,v} = k| \textbf{x}_i) = \frac{\exp \left( \tilde{\textbf{h}}_i^T \textbf{w}^{v,k}  + b^{v,k} \right)} {\sum_l \exp \left( \tilde{\textbf{h}}_i^T \textbf{w}^{v,l}  + b^{v,l} \right)} \label{eq:multiset-dist}
\end{align}
where the weights $\textbf{w}$ and biases $b$ are specific to $v$ and the multiplicity $k$.
In contrast to standard sequence-to-sequence models, this softmax is not normalized over the vocabulary but over the multiplicity, and we have distinct distributions for every vocabulary item $v$. Despite the independence assumptions in \cref{eq:multiset}, the model can still be strong because $\textbf{h}_i$ takes the entire input $\textbf{x}$ into account.

\paragraph{Training} The probability of generating the overall multiset $\textbf{m}$ as the union of all $\textbf{z}_i$ is the probability that for every vocabulary item $v$, the total number of occurrences of $v$ across all input positions $i$ sums to $\textbf{m}_v$:
\begin{align*}
    P(\textbf{m}| \textbf{x}) = \prod_v P(\textbf{z}_{1,v} + \ldots + \textbf{z}_{n,v} = \textbf{m}_v| \textbf{x})
\end{align*}
This can be computed recursively:
\begin{align*}
& P(\textbf{z}_{1,v} + \ldots + \textbf{z}_{n,v} = \textbf{m}_v\mid \textbf{x})  = \\ \nonumber 
& \sum_k P(\textbf{z}_{1,v}=k| \textbf{x}) P(\textbf{z}_{2,v} + \ldots \textbf{z}_{n,v} = \textbf{m}_v-k| \textbf{x})
\end{align*}
Let $\textbf{m}(\textbf{y})$ be the multiset of tokens in the gold sequence $\textbf{y}$. We train our model with gradient ascent to maximize the marginal log-likelihood of $\textbf{m}(\textbf{y})$:
\begin{align}
    \sum_{(\textbf{x},\textbf{y}) \in \mathcal{D}} \log P(\textbf{m}(\textbf{y})\mid \textbf{x}) \label{eq:multiset-objective}
\end{align}
Like \citet{fertility-paper}, we found it helpful to initialize the training of our model with high-confidence alignments from an IBM-1 model \citep{brown-etal-1993-mathematics} (see \cref{sec:init-multiset-model} for details).

\paragraph{Preparation for permutation}
The scoring function of the permutation model expects a sequence as input. There is no a priori obvious order for the elements within the individual multisets $\textbf{z}_i$. We handle this by imposing a canonical order $\textsc{Order}(\textbf{z}_i)$ on the elements of $\textbf{z}_i$ by sorting the multiset tokens by their id in the vocabulary. They are then concatenated to form the input $\textbf{z}' = \textsc{Order}(\textbf{z}_1) \ldots \textsc{Order}(\textbf{z}_n)$ to the permutation model.

\section{Relaxed Permutations}
\label{sec:relaxed-perms}
After predicting a multiset for every input token and arranging the elements within each multiset to form a sequence $z'$, we predict a permutation of $z'$.

We represent a permutation as a matrix $\textbf{V}$ that contains exactly one 1 in every row and column and zeros otherwise. We write $\textbf{V}_{i,j}=1$ if position $i$ in $z'$ is mapped to position $j$ in the output $y$. Let $\Perm$ be the set of all permutation matrices.

Now we formalize the parametrization of permutations as discussed in \cref{sec:setup}. 
We associate a score predicted by our neural network with each permutation $\textbf{V}$ and search for the permutation with the highest score. The score of a permutation decomposes into a sum of scores for binary `features' of $\textbf{V}$. We use two types of features.

The first type of feature is active if the permutation $\textbf{V}$ maps input position $i$ to output position $j$ (\ie $\textbf{V}_{ij}=1$). We associate this feature with the score $s_{i \mapsto j}$ and use these scores only to model what the first and the last token in the output should be. That is, $s_{i \mapsto 1}$ models the preference to map position $i$ in the input to the first position in the output, and analogously $s_{i \mapsto n}$ models the preference to put $i$ into the last position in the output. For all output positions $j$ that are neither the first nor the last position, we simply set $s_{i \mapsto j} = 0$.

The second type of feature models the jumps we introduced in \cref{sec:setup}. We introduce a feature that is $1$ iff $\textbf{V}$ contains a jump from $k$ to $i$, and associate this with a score $\sscore{k}{i}$. In order for there to be a jump from $k$ to $i$, the permutation $\textbf{V}$ must map input $i$ to some output position $j$ ($\textbf{V}_{ij}=1$) \textit{and} it must also map input position $k$ to output position $j-1$ ($\textbf{V}_{k,j-1} =1$). Hence, the product $\textbf{V}_{k,j-1} \textbf{V}_{ij}$ is 1 iff there is a jump from $k$ to $i$ at output position $j$. Based on this, the sum $\sum_j \textbf{V}_{k,j-1} \textbf{V}_{ij}$ equals 1 if there is \textit{any} output position $j$ at which there is a jump from $k$ to $i$ and 0 otherwise. This constitutes the second type of feature.

Multiplying the features with their respective scores, we want to find the highest-scoring permutation under the following overall scoring function:
\begin{align}
\argmax_{\textbf{V} \in \Perm}  & \sum_{i,j} \textbf{V}_{ij} s_{i \mapsto j} \ + \nonumber \\
 &  \sum_{i,k} \sscore{k}{i} \left( \sum_j \textbf{V}_{k,j-1} \textbf{V}_{ij}  \right) \label{eq:quadratic-assignment}
\end{align}

Let $\textbf{V}^*(s)$ be the solution to \cref{eq:quadratic-assignment} as a function of the scores. Unfortunately, $\textbf{V}^*(s)$ does not have sensible derivatives because $\Perm$ is discrete. This makes $\textbf{V}^*(s)$ unsuitable as a neural network layer. In addition, \cref{eq:quadratic-assignment} is NP-hard (see \cref{sec:np-complete}). 

We now formulate an optimization problem that approximates \cref{eq:quadratic-assignment} and which has useful derivatives.
Firstly, we relax the permutation matrix $\textbf{V}$ to a bistochastic matrix $\textbf{U}$, \ie $\textbf{U}$ has non-negative elements and every row and every column each sum to 1.
Secondly, note that \cref{eq:quadratic-assignment} contains quadratic terms. As we will discuss in the next section, our solver assumes a linear objective, so we replace $\textbf{V}_{k,j-1} \textbf{V}_{ij}$ with an auxiliary variable $\textbf{W}_{ijk}$. The variable $\textbf{W}_{ijk}$ is designed to take the value $1$ if and only if $\textbf{U}_{i,j} = 1$ \textit{and} $\textbf{U}_{k, j-1}=1$. We achieve this by coupling $\textbf{W}$ and $\textbf{U}$ using constraints. 
Then, the optimization problem becomes:
\begin{subequations} \label{eq:linear-program}
\begin{align}
    \argmax_{\textbf{U},\textbf{W}}& \sum_{i,j} \textbf{U}_{ij} s_{i \mapsto j}  + \sum_{i,j,k} \textbf{W}_{ijk} \sscore{k}{i} \tag{\ref*{eq:linear-program}} \\
    \mathrm{subject\ to\ } &\sum_i \textbf{U}_{ij}=1 \quad \quad \quad \ \forall j\label{eq:col-norm} \\
    &\sum_j \textbf{U}_{ij}=1 \quad \quad \quad \ \forall i \label{eq:row-norm} \\
    &\sum_k \textbf{W}_{ijk}= \textbf{U}_{ij} \quad \ \ \  \forall j > 1, i  \label{eq:sum-k} \\
    & \sum_i \textbf{W}_{ijk}= \textbf{U}_{k(j-1)} \: \forall j > 1, k \label{eq:sum-i} \\
    & \textbf{U},\textbf{W} \geq 0 \label{eq:non-negative}
\end{align}
\end{subequations}
Finally, in combination with the linear objective, the argmax operation still causes the solution $\textbf{U}^*(s)$ of \cref{eq:linear-program} as a function of $s$ to have no useful derivatives. This is because an infinitesimal change in $s$ has no effect on the solution $\textbf{U}^*(s)$ for almost all $s$.

To address this, we add an entropy regularization term $\tau \left( H(\textbf{U}) + H(\textbf{W})\right)$ to the objective \cref{eq:linear-program}, where $H(\textbf{U}) = -\sum_{ij} \textbf{U}_{ij} (\log \textbf{U}_{ij} - 1)$, and $\tau > 0$ determines the strength of the regularization. The entropy regularization `smooths' the solution $\textbf{U}^*(s)$ in an analogous way to softmax being a smoothed version of argmax. The parameter $\tau$ behaves analogously to the softmax temperature: the smaller $\tau$, the sharper $\textbf{U}^*(s)$ will be.
We discuss how to solve the regularized linear program in \cref{sec:bregman-for-perm}.

\paragraph{Predicting permutations} At test time, we want to find the highest scoring permutation, i.e. we want to solve \cref{eq:quadratic-assignment}. We approximate this by using $\textbf{U}^*(s)$ instead, the solution to the entropy regularized version of \cref{eq:linear-program}. Despite using a low temperature $\tau$, there is no guarantee that $\textbf{U}^*(s)$ can be trivially rounded to a permutation matrix. Therefore, we solve the linear assignment problem with $\textbf{U}^*(s)$ as scores using the Hungarian Algorithm \citep{kuhn1955hungarian}. The linear assignment problem asks for the permutation matrix $\textbf{V}$ that maximizes $\sum_{ij} \textbf{V}_{ij} \textbf{U}^*(s)_{ij}$.

\subsection{Parametrization}
We now describe how we parameterize the scores $s$ to permute the tokens into the right order. 
We first encode the original input utterance $x$ like in \cref{eq:encode} to obtain a hidden representation $\textbf{h}_i$ for input token $\textbf{x}_i$. %
Let $a$ be the function that maps $a(i) \mapsto j$ if the token in position $i$ in $z'$ came from the multiset that was generated by token $\textbf{x}_j$. For example, in \cref{fig:intro}, $a(6)=3$ since \texttt{sleep} was predicted from input token \textit{slept}. We then define the hidden representation $\textbf{h}'_i$ as the concatenation of $\textbf{h}_{a(i)}$ and an embedding of $z'_i$: 
\begin{align}
    \textbf{h}'_i = \left[ \textbf{h}_{a(i)}; \textsc{embed}(z'_i) \right] \label{eq:hidden-rep-perm}
\end{align}
We parameterize the scores for starting the output with token $i$ as $$s_{i \mapsto 1} = \textbf{w}_{\text{start}}^T \textsc{ff}_{\text{start}}(\textbf{h}'_i)$$ and analogously for ending it with token $i$: $$s_{i \mapsto n} = \textbf{w}_{\text{end}}^T \textsc{ff}_{\text{end}}(\textbf{h}'_i)  $$
We set $s_{i \mapsto j} = 0$ for all other $i, j$.

We parameterize the jump scores $\sscore{k}{i}$ using Geometric Attention \citep{csordas2022} from $\textbf{h}'_k$ to $\textbf{h}'_i$. 
Intuitively, Geometric Attention favours selecting the `matching' element $\textbf{h}'_i$ that is closest to $\textbf{h}'_k$ in terms of distance $|i-k|$ in the string. We refer to \citet{csordas2022} for details.

\subsection{Learning Permutations}

\label{sec:learn-permutations}
We now turn to training the permutation model. At training time, we have access to the gold output $\textbf{y}$ and a sequence $\textbf{z}'$ from the output of the multiset tagging (see the end of \cref{sec:multisets}). We note that whenever $\textbf{y}$ (or $\textbf{z}'$) contains one vocabulary item at least twice, there are multiple permutations that can be applied to $\textbf{z}'$ to yield $\textbf{y}$. Many of these permutations will give the right result for the wrong reasons
and the permutation that is desirable for generalization is latent. 
\begin{figure}[t]
    \centering
    \includegraphics[width=\linewidth]{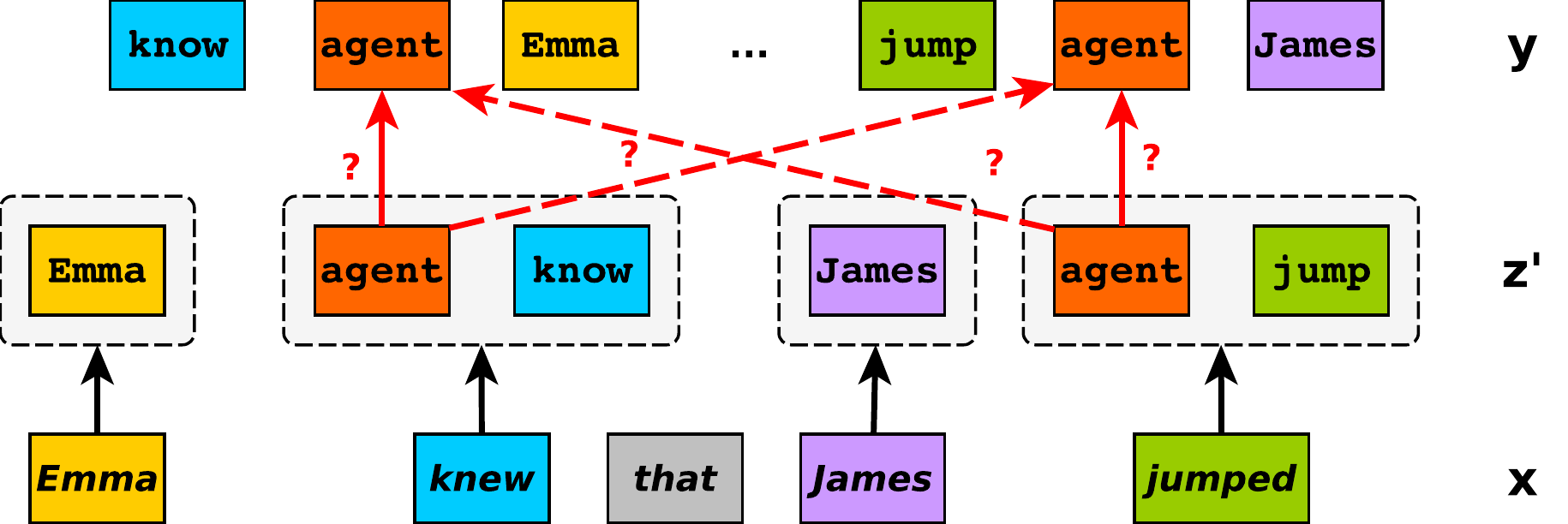}
    \caption{Two possible permutations (represented with red arrows) of $\textbf{z}'$ that yield the same result $\textbf{y}$. Only the permutation marked with solid arrows corresponds to the right linguistic relation that will generalize well.}
    \label{fig:two-perm-agent}
\end{figure}
For example, consider \cref{fig:two-perm-agent}. The token \texttt{agent} is followed by the entity who performs the action, whereas \texttt{theme} is followed by the one affected by the action.
The permutation indicated by dashed arrows generalizes poorly to a sentence like \textit{Emma knew that James slid} since \textit{slid} will introduce a \texttt{theme} role rather than an \texttt{agent} role (as the sliding is happening to James). Thus, this permutation would then lead to the incorrect output \texttt{know \underline{theme} Emma ... slide \underline{agent} James}, in which Emma is affected by the knowing event and James is the one who slides something.

In order to train the permutation model in this setting, we use a method that is similar to EM for structured prediction.\footnote{It can also be derived as a lower bound to the marginal log-likelihood $\log P_{\theta}(\textbf{y}|\textbf{x},\textbf{z}')$ (see \cref{sec:loss-derivation}), and be seen as a form of posterior regularization \citep{ganchev2010posterior}}
During training, the model output $\textbf{U}^*(s)$ and $\textbf{W}^*(s)$ often represents a soft permutation that does \textit{not} permute $\textbf{z}'$ into $\textbf{y}$. Our goal is to push the model output into the space of (soft) permutations that lead to $\textbf{y}$. 
More formally, let $\textbf{Q} \in \mathcal{Q}(\textbf{y},\textbf{z}')$ be the set of bistochastic matrices such that $\textbf{Q}_{i,j} = 0$ iff $\textbf{z}'_i \neq \textbf{y}_j$.
That is, any permutation included in $\mathcal{Q}(\textbf{y},\textbf{z}')$ leads to the gold output $\textbf{y}$ when applied to $\textbf{z}'$.

First, we project the current prediction $\textbf{U}^*(s)$ and $\textbf{W}^*(s)$ into $\mathcal{Q}(\textbf{y},\textbf{z}')$ using the KL divergence as a measure of distance (E-step):
\begin{align}
    \hat{\textbf{U}}, \hat{\textbf{W}} = \argmin_{\textbf{U} \in \mathcal{Q}(\textbf{y}, \textbf{z}'),\textbf{W}} & \kl (\textbf{U} \dbar \textbf{U}^*(s)) + \label{eq:min-cross-entropy} \\
     & \kl (\textbf{W} \dbar \textbf{W}^*(s)) \nonumber
\end{align}
subject to \cref{eq:col-norm} to  \cref{eq:non-negative}.

Similar to the M-step of EM, we then treat $\hat{\textbf{U}}$ and $\hat{\textbf{W}}$ as soft gold labels and train our model to minimize the KL divergence between labels and model:
\begin{align*}
    \kl(\hat{\textbf{U}} \dbar \textbf{U}^*(s)) + \kl(\hat{\textbf{W}} \dbar \textbf{W}^*(s))
\end{align*}
\cref{eq:min-cross-entropy} can be solved in the same way as the entropy regularized version of \cref{eq:linear-program} because expanding the definition of KL-divergence leads to a regularized linear program with a similar feasible region (see \cref{sec:argmax-and-kl} for details).

\section{Inference for Relaxed Permutations}
\label{sec:bregman-for-perm}

Now we describe how to solve the entropy regularized form of \cref{eq:linear-program} and how to backpropagate through it. This section may be skipped on the first reading as it is not required to understand the experiments; we note that the resulting algorithm (\cref{alg:kl-project}) is conceptually relatively simple. Before describing our method, we explain the general principle.

\subsection{Bregman's Method}
\label{sec:bregman}
Bregman's method \citep{bregman1967relaxation} is a method for constrained convex optimization. In particular, it can be used to solve problems of the form
\begin{align}
    \textbf{x}^* = \argmax_{\textbf{x}\in C_0 \cap C_1 \ldots \cap C_{n}, \textbf{x} \geq \textbf{0} } \textbf{s}^T \textbf{x} + \underbrace{\tau H(\textbf{x})}_{\text{regularizer}} \label{eq:background-bregman}
\end{align}
where $C_0, \ldots, C_n$ are linear equality constraints, $H(\textbf{x}) = -\sum_i \textbf{x}_i (\log \textbf{x}_i - 1)$ is a form of entropy regularization, and $\tau$ determines the strength of the regularization. Note that our parameterization of permutations (\cref{eq:linear-program}) has this form. 

Bregman's method is a simple iterative process. We start with the scores $\textbf{s}$ and then cyclically iterate over the constraints and project the current estimate $\textbf{x}^i$ onto the chosen constraint until convergence:
\begin{equation}\label{eq:bregmans-method}
    \begin{aligned}
    \textbf{x}^0 &= \exp \frac{\textbf{s}}{\tau} \\
    \textbf{x}^{i+1} &= \argmin_{\textbf{x} \in C_{i \text{ mod } (n-1)}} \kl(\textbf{x} \dbar \textbf{x}^i) 
\end{aligned}
\end{equation}
where $\kl(\textbf{x}\mid \textbf{y}) = \sum_i \textbf{x}_i \log \frac{\textbf{x}_i}{\textbf{y}_i} - \textbf{x}_i + \textbf{y}_i$ is the generalized KL divergence. We call $\argmin_{\textbf{x} \in C} \kl(\textbf{x} \mid \textbf{x}^i) $ a KL projection. In order to apply Bregman's method, we need to be able to compute the KL projection $\argmin_{\textbf{x} \in C} \kl(\textbf{x} \mid \textbf{x}^i)$ for \textit{all} $C_0, \ldots, C_n$ in closed-form. We discuss how to do this for \cref{eq:linear-program} in the next section.

As an example, consider a problem of the form \cref{eq:background-bregman} with a single linear constraint $C^{\Delta} = \{ \textbf{x} \mid  \sum_i \textbf{x}_i = 1  \}$. In this case, Bregman's method coincides with the softmax function. This is because the KL projection $\textbf{x}^* = \argmin_{\textbf{x} \in C^{\Delta}} \kl(\textbf{x}\dbar\textbf{y})$ for $\textbf{y} > \textbf{0}$ has the closed-form solution $\textbf{x}^*_i = \frac{\textbf{y}_i}{\sum_i \textbf{y}_i}$.

If we have a closed-form expression for a KL projection (such as normalizing a vector), we can use automatic differentiation to backpropagate through it. 
To backpropagate through the entire solver, we apply automatic differentiation to the composition of all projection steps.

\subsection{Bregman's Method for \cref{eq:linear-program}}
In order to apply Bregmans' method to solve the entropy regularized version of \cref{eq:linear-program}, we need to decompose the constraints into sets which we can efficiently project onto.
We choose the following three sets here: (i), containing \cref{eq:col-norm} and \cref{eq:sum-k}, and (ii), containing \cref{eq:col-norm} and \cref{eq:sum-i}, and finally (iii), containing only \cref{eq:row-norm}.
We now need to establish what the KL projections are for our chosen sets. For (iii), the projection is simple:

\algrenewcommand\algorithmicindent{0.51em}%
\begin{algorithm}[t]
	\caption{Bregman's method for \cref{eq:linear-program} with entropic regularization}
	
	\begin{algorithmic}
		\Function{bregman}{s, $\tau$}
		\State $\textbf{U}_{ij} =\exp(\tau^{-1} s_{i \mapsto j}$)
		\State $\textbf{W}_{ijk} = \exp(\tau^{-1} \sscore{k}{i})$
		\While{within budget and not converged}
		\State $\textbf{U}, \textbf{W} = $ KL project($\textbf{U},\textbf{W}$; \ref{eq:col-norm},\ref{eq:sum-k}) with Prop.~\ref{th:kl-project}
		\State $\textbf{U}, \textbf{W} = $ KL project($\textbf{U},\textbf{W}$; \ref{eq:col-norm},\ref{eq:sum-i}) with Prop.~\ref{th:kl-project}
		\State $\textbf{U}= $ KL project ($\textbf{U}$, \ref{eq:row-norm}) with Prop.~\ref{th:kl-project-simplex}
		\EndWhile
		\State \textbf{return} $\textbf{U}, \textbf{W}$
		\EndFunction
	\end{algorithmic}
	\label{alg:kl-project}
\end{algorithm}

\begin{prop}
(\citet{benamou}, Prop.~1) For $\textbf{A}, \textbf{m} > 0$, the KL projection $\argmin_{\textbf{U}} \kl(\textbf{U} \dbar \textbf{A})$ subject to $\sum_j \textbf{U}_{ij} = \textbf{m}_i$ is given by $\textbf{U}^*_{ij} = \textbf{m}_i \frac{\textbf{A}_{ij}}{\sum_{j'} \textbf{A}_{ij'}}$.
\label{th:kl-project-simplex}
\end{prop}

Let us now turn to (i) and (ii). The constraints \cref{eq:sum-i} and \cref{eq:sum-k} are structurally essentially the same, meaning that we can project onto (ii) in basically the same manner as onto (i). 
We project onto (i), with the following proposition:
\begin{prop}
For $\textbf{A},\textbf{B} > \textbf{0}$, the KL projection
\begin{argmini}
{\textbf{U},\textbf{W}}{\kl(\textbf{U} \dbar \textbf{A}) + \kl(\textbf{W} \dbar \textbf{B})}{}{}
\addConstraint{\sum_i \textbf{U}_{ij}}{=1& \forall j  }
\addConstraint{\sum_k \textbf{W}_{ijk}}{ = \textbf{U}_{ij}& \: \: \forall j, i}
\end{argmini}
is given by:
\begin{align*}
    \textbf{U}^*_{ij} &= \frac{\textbf{T}_{ij}}{\sum_{i'} \textbf{T}_{i'j}} \\
    \textbf{W}^*_{ijk} &= \textbf{U}^*_{ij} \frac{\textbf{B}_{ijk}}{\sum_{k'} \textbf{B}_{ijk'}}
\end{align*}
where $\textbf{T}_{ij} = \sqrt{\textbf{A}_{ij} \cdot \sum_k \textbf{B}_{ijk}}$.
\label{th:kl-project}
\end{prop}
The proof can be found in \cref{sec:proof-kl-projection}.
We present the overall algorithm in \cref{alg:kl-project}, and note that it is easy to implement for GPUs.
In practice, we implement all projections in log space for numerical stability.

\section{Evaluation}

\subsection{Doubling Task}
\label{sec:doubling}
\begin{figure}[t]
    \centering
    \includegraphics[width=\linewidth]{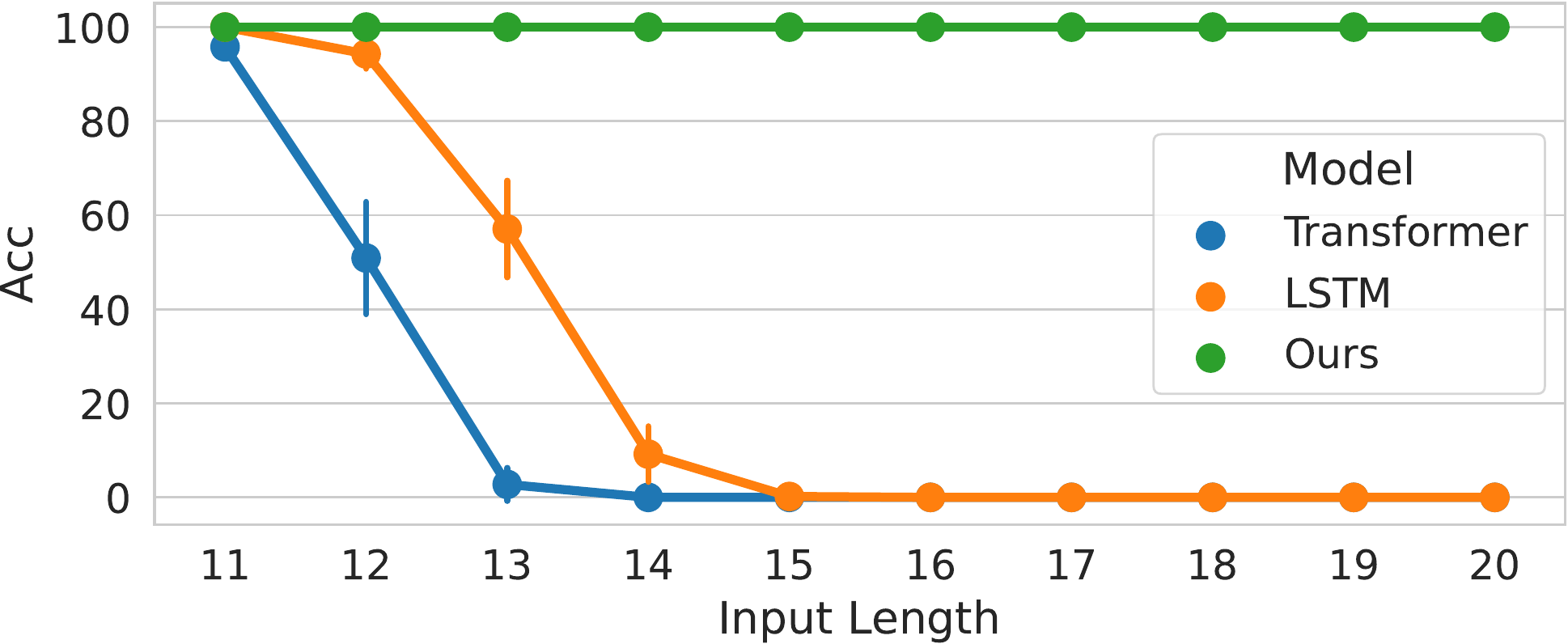}
    \caption{Accuracy by input length for doubling task.}
    \label{fig:doubling-plot}
\end{figure}

Our permutation model is very expressive and is not limited to synchronous context-free languages. This is in contrast to the formalisms that other approaches rely on  \citep{wang2021structured,fertility-paper}. To evaluate if our model can structurally generalize beyond the synchronous context-free languages in practice, we consider the function $F = \{(w,ww) \mid w \in \Sigma^*\}$. This function is related to processing challenging natural language phenomena such as reduplication and cross-serial dependencies.
We compare our model with an LSTM-based seq2seq model with attention and a Transformer in the style of \citet{csordas-etal-2021-devil} that uses a relative positional encoding. Since the input is a sequence of symbols rather than English, we replace RoBERTa with a bidirectional LSTM and use randomly initialized embeddings. The models are trained on inputs of lengths 5 to 10 and evaluated on longer examples.
The results can be found in \cref{fig:doubling-plot}. All models get perfect or close to perfect accuracy on inputs of length 11 but accuracy quickly deteriorates for the LSTM and the Transformer. In contrast, our model extrapolates very well to longer sequences.

\subsection{COGS}
\label{sec:cogs}
COGS is a synthetic benchmark for compositional generalization introduced by \citet{kim-linzen-2020-cogs}. Models are tested for 21 different cases of generalization, 18 of which focus on using a lexical item in new contexts (\textbf{Lex}). There are 1000 instances per generalization case. Seq2seq models struggle in particular with the \textit{structural} generalization tasks \citep{yao-koller-2022}, and we focus on those: (i) generalization to deeper \textbf{PP} recursion than seen during training ("Emma saw a hedgehog on a chair in the garden beside the tree..."), (ii) deeper \textbf{CP} recursion ("Olivia mentioned that James saw that Emma admired that the dog slept"), and (iii) PPs modifying subjects when PPs modified only objects in the training data (\textbf{OS}).

We follow previous work and use a lexicon \citep{akyurek-andreas-2021-lexicon} to map some input tokens to output tokens (see \cref{sec:lexicon} for details). We also use this mechanism to handle the variable symbols in COGS.

\begin{table}[t]
\resizebox{\linewidth}{!}{
\begin{tabular}{llllll}
\toprule
\textbf{Model} &\textbf{OS} & \textbf{CP} & \textbf{PP} & \textbf{Lex} & \textbf{Total} \\
\midrule
Q'22\pretrained\treemodel  & - & - & - & - & \textbf{99} \\
D'22\pretrained\treemodel & - & - & - & - & \textbf{99}\\ %
LexLSTM & 0 & 0 & 1 & 96 & 82 \\
Dangle & 0 & 14 & 14 & 98 & 85 \\
Dangle\pretrained & 0 & 25 & 35 & \textbf{99} & 88 \\
T5-base\pretrained & 0 & 13 & 18 & \textbf{99} & 86 \\
\midrule
Ours\pretrained & \textbf{9}\ci{13} & \textbf{79}\ci{11} & \textbf{85}\ci{14} & 97\ci{2} & 92\ci{2} \\
Ours\pretrained - Lex & 5\ci{6} & 71\ci{26} & 82\ci{20} & 79\ci{1} & 75\ci{2} \\
\bottomrule
\end{tabular}}
\caption{Exact match accuracy on \textbf{COGS} by generalization type. \pretrained refers to models using pretrained transformers, and \treemodel refers to models implicitly or explicitly using trees. Q'22 is \citet{qiu-etal-2022-improving}, D'22 is \citet{drozdov2022compositional}. LexLSTM is the model by \citet{akyurek-andreas-2021-lexicon} and Dangle is the model by \citet{zheng-lapata-2021-compositional-generalization}. We report their results for T5.}
\label{tab:cogs-exact}
\vspace{12pt}
\resizebox{\linewidth}{!}{
\begin{tabular}{llllll}
\toprule
\textbf{Model} &\textbf{OS} & \textbf{CP} & \textbf{PP} & \textbf{Lex} & \textbf{Total} \\
\midrule
Lear\treemodel & \textbf{93} & \textbf{100} & \textbf{99} & \textbf{99} & \textbf{99} \\
AM parser\pretrained\treemodel & 72 & \textbf{100} & 97 & 76 & 78 \\
Dangle & 8 & 14 & 14 & \textbf{99} & 87 \\
\midrule
Ours\pretrained & 33\ci{24} & 82\ci{11} & 91\ci{5} & 97\ci{2} & 93\ci{2} \\
Ours\pretrained - Lex & 35\ci{22} & 73\ci{27} & 92\ci{5} & 79\ci{1} & 77\ci{2} \\
\bottomrule
\end{tabular}}
\caption{Accuracy on \textbf{COGS} when the order of conjuncts is disregarded or established in post-processing. For Lear \citep{liu-etal-2021-learning-algebraic} and the AM parser \citep{groschwitz-etal-2018-amr,weissenhorn-etal-2022-compositional}, we report numbers of \citet{weissenhorn-etal-2022-compositional}.}
\label{tab:cogs-order-invariant}
\vspace{-11pt}
\end{table}

We report the means and standard deviations for 10 random seeds in \cref{tab:cogs-exact}.
Our approach obtains high accuracy on CP and PP recursion but exact match accuracy is low for OS. This is in part because our model sometimes predicts semantic representations for OS that are equivalent to the gold standard but use a different order for the conjuncts. Therefore, we report accuracy that accounts for this in \cref{tab:cogs-order-invariant}. In both tables, we also report the impact of using a simple copy mechanism instead of the more complex lexicon induction mechanism (-Lex). Our model outperforms all other non-tree-based models by a considerable margin.

\paragraph{Structural generalization without trees} All previous methods that obtain high accuracy on recursion generalization on COGS use trees. Some approaches directly predict a tree over the input \citep{liu-etal-2021-learning-algebraic, weissenhorn-etal-2022-compositional}, while others use derivations from a grammar for data augmentation \citep{qiu-etal-2022-improving} or decompose the input along a task-specific parse tree \citep{drozdov2022compositional}. 
Our results show that trees are not as important for compositional generalization as their success in the literature may suggest, and that weaker structural assumptions already reap some of the benefits. 

\paragraph{Logical forms with variables} COGS uses logical forms with variables, which were removed in conversion to variable-free formats for evaluation of some approaches \citep{zheng-lapata-2021-compositional-generalization, qiu-etal-2022-improving,drozdov2022compositional}. Recently, \citet{wu2023recogs} have argued for keeping the variable symbols because they are important for some semantic distinctions; we keep the variables.

\subsection{ATIS}
\label{sec:atis}

\begin{table}[t]
\centering
\begin{tabular}{lll}
\toprule
\textbf{Model} & \textbf{iid} & \textbf{Length} \\
\midrule
LSTM seq2seq\nogrammar & 75.98\cinf{1.30} & 4.95\cinf{2.16} \\
Transformer\nogrammar & 75.76\cinf{1.43} & 1.15\cinf{1.41} \\
BART-base\nogrammar\pretrained & \textbf{86.96}\cinf{1.26} & 19.03\cinf{4.57} \\
\shortfert\nogrammar\treemodel & 68.26\cinf{1.53} & 29.91\cinf{2.91} \\
\shortfert\grammar\treemodel & 74.15\cinf{1.35} & 35.41\cinf{4.09} \\
\midrule
Ours\nogrammar \pretrained & 76.65\cinf{1.67} & \textbf{41.39}\cinf{13.47} \\
Ours\nogrammar &73.93\cinf{1.43} & 38.79\cinf{7.11} \\
\bottomrule
\end{tabular}
\caption{Accuracy on \textbf{ATIS}. \grammar indicates grammar-based decoding. \shortfert is the model of \citet{fertility-paper}.} %
    \label{tab:atis}
\end{table}

While COGS is a good benchmark for compositional generalization, the data is synthetic and does not contain many phenomena that are frequent in semantic parsing on real data, such as paraphrases that map to the same logical form. ATIS \citep{dahl-etal-1994-expanding} is a realistic English semantic parsing dataset with executable logical forms. 
\ifreviewmode
 We follow the setup of \citet{fertility-paper} (\shortfert) and use the variable-free FunQL representation \citep{guo-etal-2020-benchmarking-meaning}.
\else
 We follow the setup of our previous work \citep{fertility-paper} (\shortfert) and use the variable-free FunQL representation \citep{guo-etal-2020-benchmarking-meaning}.
\fi
Apart from the usual iid split, we evaluate on a length split, where a model is trained on examples with few conjuncts and has to generalize to longer logical forms with more conjuncts. For a fair comparison with previous work, we do not use the lexicon/copying. We also evaluate a version of our model without RoBERTa that uses a bidirectional LSTM and GloVe embeddings instead. This mirrors the model of \shortfert.

\cref{tab:atis} shows mean accuracy and standard deviations over 5 runs. Our model is competitive with non-pretrained models in-distribution, and outperforms all other models on the length generalization. The high standard deviation on the length split stems from an outlier run with 18\% accuracy -- the second worst-performing run achieved an accuracy of 44\%. %
Even without pretraining, our model performs very well. In particular, without grammar-based decoding our model performs on par or outperforms \shortfert \textit{with} grammar-based decoding. 

The runtime of the model in \shortfert is dominated by the permutation model and it takes up to 12 hours to train on ATIS. 
\ifreviewmode
Training our model presented only takes around 2 hours for both stages.
\else
Training the model presented here only takes around 2 hours for both stages.
\fi

\begin{table}[t]
    \centering
    \resizebox{\linewidth}{!}{
\begin{tabular}{lllll}
\toprule
& & \textbf{Freq} & \textbf{Seq} & \textbf{Seq/Freq} \\
\midrule
\multirow{3}{*}{ COGS } & OS & 50\ci{34} & 9\ci{13} & 11\ci{15} \\
& CP & 97\ci{5} & 79\ci{11} & 82\ci{13} \\
& PP & 99\ci{0} & 85\ci{14} & 85\ci{15} \\
\midrule
\multirow{2}{*}{ ATIS } & iid & 77.6\ci{1.4} & 76.7\ci{1.7} & 98.7\ci{0.5} \\
& Length & 42.2\ci{13.6} & 41.4\ci{13.5} & 97.8\ci{0.8} \\
\bottomrule
\end{tabular}}
    \caption{Performance breakdown of the first and second stage. `Freq' refers to accuracy measured on the predicted multiset; it measures performance of the first stage. `Seq' measures the accuracy of both stages. `Seq/Freq' is the percentage of correct predictions given that the multiset is predicted correctly.}
    \label{tab:error-break-down-full}
\end{table}
\paragraph{Performance breakdown} In order for our approach to be accurate, both the multiset tagging model and the permutation model have to be accurate. \cref{tab:error-break-down-full} explores which model acts as the bottleneck in terms of accuracy on ATIS and COGS. The answer depends on the dataset: for the synthetic COGS dataset, predicting the multisets correctly is easy except for OS, and the model struggles more with getting the permutation right. In contrast, for ATIS, the vast majority of errors can be attributed to the first stage.

\subsection{Okapi}
\label{sec:okapi}

\begin{table}[t]
    \centering
    \begin{tabular}{lrrr}
    \toprule
    \textbf{Model} & \textbf{Calendar} & \textbf{Doc} & \textbf{Email} \\
    \midrule
BART-base\nogrammar\pretrained & 36.7\cinf{3.0} & 0.6\cinf{0.3} & 20.5\cinf{9.8} \\
\shortfert\nogrammar\treemodel & 57.2\cinf{19.9} & 36.1\cinf{5.6} & 43.9\cinf{3.8} \\
\shortfert\grammar\treemodel & 69.5\cinf{13.9} & 42.4\cinf{5.7} & 55.6\cinf{2.7} \\
\midrule 
Ours\pretrained & \textbf{74.3}\cinf{3.5} &  \textbf{57.8}\cinf{5.5} & \textbf{60.6}\cinf{4.8} \\
Ours & 65.6\cinf{2.8} & 41.4\cinf{4.9} & 47.6\cinf{4.5} \\
    \bottomrule
    \end{tabular}
    \caption{Accuracy on length splits by domain on \textbf{Okapi}.}
    \label{tab:okapi}
\end{table}

Finally, we consider the recent Okapi \citep{okapi} semantic parsing dataset, in which an English utterance from one of three domains (Calendar, Document, Email) has to be mapped to an API request. We again follow the setup of \shortfert and evaluate on their length split, where a model has to generalize to longer logical forms. In contrast to all other datasets we consider, Okapi is quite noisy because it was collected with crowd workers. This presents a realistic additional challenge on top of the challenge of structural generalization.

The results of 5 runs can be found in \cref{tab:okapi}. Our model outperforms both BART \citep{lewis-etal-2020-bart} and the model of \shortfert. In the comparison without pretraining, our model also consistently achieves higher accuracy than the comparable model of \shortfert without grammar-based decoding.

\section{Related Work}
\paragraph{Predicting permutations}
\citet{mena2018learning} and \citet{lyu-titov-2018-amr} use variational autoencoders based on the Sinkhorn algorithm to learn latent permutations. The Sinkhorn algorithm \citep{sinkhorn} is also an instance of Bregman's method and solves the entropy regularized version of \cref{eq:linear-program} without the $\textbf{W}$-term. This parameterization is considerably weaker than ours since it cannot capture our notion of `jumps'.

\citet{wang2021structured} compute soft permutations as an expected value by marginalizing over the permutations representable by ITGs \citep{wu-1997-stochastic}. This approach is exact but excludes some permutations. In particular, it excludes permutations needed for COGS.\footnote{This is mostly due to the treatment of definite descriptions, which appear clustered together at the start of the logical form.} In addition, the algorithm they describe takes a lot of resources as it is both $O(n^5)$ in memory and compute.
\citet{devatine-etal-2022-ordonnancement} investigate sentence reordering methods. They use bigram scores, which results in a similar computational problem to ours. However, they deal with it by restricting what permutations are possible to enable tractable dynamic programs.  %
\citet{eisner2006local} propose local search methods for decoding permutations for machine translation.  

Outside of NLP, \citet{kushinsky-2019} have applied Bregman's method to the quadratic assignment problem, which \cref{eq:quadratic-assignment} is a special case of. Since they solve a more general problem, using their approach for \cref{eq:linear-program} would require $O(n^4)$ rather than $O(n^3)$ variables in the linear program.

\paragraph{Compositional generalization}
Much research on compositional generalization has focused on lexical generalization with notable success \citep{andreas-2020-good, akyurek-andreas-2021-lexicon, conklin-etal-2021-meta, csordas-etal-2021-devil}. Structural generalization remains more challenging for seq2seq models \citep{yao-koller-2022}.

\citet{zheng-lapata-2022-disentangled} modify the transformer architecture and re-encode the input and partially generated output for every decoding step to disentangle the information in the representations. 
Structure has also been introduced in models by means of grammars: \citet{qiu-etal-2022-improving} heuristically induce a quasi-synchronous grammar (QCFG, \citet{smith-eisner-2006-quasi}) and use it for data augmentation for a seq2seq model. \citet{kim-2021-nqscfg} introduces neural QCFGs which perform well on compositional generalization tasks but are very compute-intensive.
Other works directly parse into trees or graphs inspired by methods from syntactic parsing \citep{liu-etal-2021-learning-algebraic, herzig-berant-2021-span, weissenhorn-etal-2022-compositional, jambor-bahdanau-2022-lagr, petit2023graph}.

Several approaches, including ours, have decoupled the presence or absence of output tokens from their order:
\citet{wang2021structured} train a model end-to-to-end to permute the input (as discussed above) and then monotonically translate it into an output sequence. \citet{fertility-paper} also present an end-to-end differentiable model that first applies a `fertility step' which predicts for every word how many copies to make of its representation, and then uses the permutation method of \citet{wang2021structured} to reorder the representation before translating them. 
\citet{cazzaro2022translate} first translate the input monotonically and feed it into a second model. They use alignments from an external aligner to train the first model. The second model is a tagger or a pretrained seq2seq model and predicts the output as a permutation of its input. We compare against such a baseline for permutations in \cref{appendix:additional-results}, finding that it does not work as well as ours in the compositional generalization setups we consider.

\section{Conclusion}
In this paper, we have presented a flexible new seq2seq model for semantic parsing. Our approach consists of two steps: We first tag each input token with a multiset of output tokens. Then we arrange those tokens into a sequence using a permutation model. 
We introduce a new method to predict and learn permutations based on a regularized linear program that does not restrict what permutations can be learned.
The model we present has a strong ability to generalize compositionally on synthetic and natural semantic parsing datasets. Our results also show that trees are not necessarily required to generalize well to deeper recursion than seen at training time.

\section*{Limitations}
The conditional independence assumptions are a limitation for the applicability of our multiset tagging model. For example, the independence assumptions are too strong to apply it to natural language generation tasks such as summarization. From a technical point of view, the independence assumptions are important to be able to induce the latent assignment of output tokens to multisets efficiently. %
Future work may design multiset tagging methods that make fewer independence assumptions.

While our method for predicting permutations is comparatively fast and only has a memory requirement of $O(n^3)$, inference on long sequences, \eg with more than 100 tokens, remains somewhat slow. In future work, we plan to investigate other approximate inference techniques like local search and dual decomposition.

Regarding the importance of trees for compositional generalization, our model has no explicit structural inductive bias towards trees. However, we do not exclude that the pretrained RoBERTa model that we use as a component \textit{implicitly} captures trees or tree-like structures to a certain degree.

\ifreviewmode

\else

\section*{Acknowledgements}
We thank Bailin Wang and Jonas Groschwitz for technical discussions; we thank Hao Zheng for discussions and for providing system outputs for further analysis. We also say thank you to Christine Sch\"afer and Agostina Calabrese for their comments on this paper.

ML is supported by the UKRI Centre for Doctoral Training in Natural Language Processing, funded by the UKRI (grant EP/S022481/1), the University of Edinburgh, School of Informatics and School of Philosophy, Psychology \& Language Sciences, and a grant from Huawei Technologies. IT is supported by the Dutch National Science Foundation (NWO Vici VI.C.212.053).

\fi

\bibliography{anthology,custom}
\bibliographystyle{acl_natbib}

\appendix

\section{Math Details}

\subsection{NP-hardness}
\label{sec:np-complete}
We show that \cref{eq:quadratic-assignment} can be used to decide the Hamiltonian Path problem. Let $G=(V,E)$ be a graph with nodes $V=\{1, 2, \ldots, n \}$. A Hamiltonian path $P = v_1, v_2, \ldots, v_n$ is a path in $G$ (i.e. $(v_i, v_{i+1}) \in E$ for all $i$) such that each node of $G$ appears exactly once. Deciding if a graph has a Hamiltonian path is NP-complete.

\paragraph{Reduction of Hamiltonian path to \cref{eq:quadratic-assignment}}
Note that a necessary but not sufficient condition for $P$ to be a Hamiltonian path is that $P$ is a permutation of $V$. This will be ensured by the constraints on the solution in \cref{eq:quadratic-assignment}.

We construct a score function
\begin{align}
    \sscore{k}{i} = \begin{cases}
    1 & \text{ if }(k,i) \in E \\
    0 & \text{ else}
    \end{cases} \label{eq:np-complete-score-function}
\end{align}
and let $s_{i \mapsto j} = 0$ for all $i, j$. If we find the solution of \cref{eq:quadratic-assignment} for the score function \cref{eq:np-complete-score-function}, we obtain a permutation $P$ of $V$, which may or may not be a path in $G$. In a path of $n$ nodes, there are $n-1$ edges that are crossed. If the score of the solution is $n-1$, then all node pairs $(v_i, v_{i+1})$ that are adjacent in $P$ must have had a score of $1$, indicating an edge $(v_i, v_{i+1}) \in E$. Therefore, $P$ must be a Hamiltonian path. If the score of the solution is less than $n-1$, then there is no permutation of $V$ that is also a path, and hence $G$ has no Hamiltonian path.

\subsection{Proof of Proposition \ref{th:kl-project}}
\label{sec:proof-kl-projection}

\begin{figure*}[t]
    \begin{align*}
    & \kl(\textbf{x}\dbar\textbf{z}) + \kl(\textbf{Y}^*(\textbf{x}) \dbar \textbf{W}) \\
    &= \kl(\textbf{x}\dbar\textbf{z})  + \sum_{i, j} \frac{\textbf{x}_i \textbf{W}_{i,j}}{\sum_{j'} \textbf{W}_{i,j'}} (\log \frac{\frac{x_i \cancel{\textbf{W}_{i,j}}}{\sum_{j'} \textbf{W}_{i,j'}}}{\cancel{\textbf{W}_{i,j}}} -1) \\
    &=   \kl(\textbf{x}\dbar\textbf{z})  + \sum_i \frac{\textbf{x}_i \cancel{\sum_{j} \textbf{W}_{i,j}}}{\cancel{\sum_{j'} \textbf{W}_{i,j'}}} (\log \frac{\textbf{x}_i}{\sum_{j'} \textbf{W}_{i,j'}} -1) \\
    &= \sum_i \textbf{x}_i ( \log (\frac{\textbf{x}_i}{\textbf{z}_i}) - 1 + \log \frac{\textbf{x}_i}{\sum_{j'} \textbf{W}_{i,j'}} -1) \\
    &= \sum_i \textbf{x}_i ( 2 \log \textbf{x}_i - \log \textbf{z}_i - \log \left( \sum_{j'} \textbf{W}_{i,j'} \right) -2) \\
    &= 2 \sum_i \textbf{x}_i ( \log \textbf{x}_i -1 - \underbrace{\frac{1}{2} ( \log \textbf{z}_i + \log \left[ \sum_{j'} \textbf{W}_{i,j'} \right])}_{\log \textbf{q}_i}) \\
    &= 2 \cdot \kl(\textbf{x}\dbar\textbf{q})
\end{align*}
\caption{Rewriting the objective function of \cref{eq:one-variable-prob}. We note that the generalized KL divergence $\kl(\textbf{x}\mid \textbf{y}) = \sum_i \textbf{x}_i \log \frac{\textbf{x}_i}{\textbf{y}_i} - \textbf{x}_i + \textbf{y}_i$ simplifies to $\kl(\textbf{x}\mid \textbf{y}) = \sum_i \textbf{x}_i (\log \frac{\textbf{x}_i}{\textbf{y}_i} -1)$ because we want to find the argmin wrt to $\textbf{x}$.}
\label{fig:rewriting-objective-function}
\end{figure*}

We now prove Prop.~\ref{th:kl-project} using a very similar technique as \citet{kushinsky-2019}. As the constraints in Prop.~\ref{th:kl-project} for any value of $j$ do not interact with constraints for other values of $j$, we can assume w.l.o.g. that $j$ takes a single value only and drop it in the notation. We want to solve:
\begin{argmini}
{\textbf{x},\textbf{Y}}{\kl(\textbf{x} \dbar \textbf{z}) + \kl(\textbf{Y} \dbar \textbf{W})}{}{}
\addConstraint{\sum_i\textbf{x}_{i}}{=1}
\addConstraint{\sum_k \textbf{Y}_{ik}}{ = \textbf{x}_{i} \: \: \forall i}
\end{argmini}
We can find $\textbf{Y}^*=\argmin_{\textbf{Y}} \kl(\textbf{Y} \dbar \textbf{W})$ subject to $\sum_j \textbf{Y}_{i,j} = \textbf{x}_i$ based on Prop.~\ref{th:kl-project-simplex}: $\textbf{Y}^*_{i,j} = \frac{\textbf{x}_i \textbf{W}_{i,j}}{\sum_j \textbf{W}_{i,j}}$. That is, we can express $\textbf{Y}^*$ as a function of $\textbf{x}$ (which we write $\textbf{Y}^*(\textbf{x})$), and therefore our overall problem is now a problem in one variable ($\textbf{x}$):
\begin{argmini}
{\textbf{x}}{\kl(\textbf{x} \dbar \textbf{z}) + \kl(\textbf{Y}^*(\textbf{x}) \dbar \textbf{W})}{}{}
\addConstraint{\sum_i \textbf{x}_{i}}{=1}
\label{eq:one-variable-prob}
\end{argmini}
We can now rewrite the objective function as 
\begin{argmini}
{\textbf{x}}{\kl(\textbf{x}\dbar\textbf{q})}{}{}
\addConstraint{\sum_i \textbf{x}_{i}}{=1}
\end{argmini}
where $\textbf{q}_i = \sqrt{\textbf{z}_i \cdot \sum_{j'} \textbf{W}_{i,j'} }$. This step is justified in detail in \cref{fig:rewriting-objective-function}.

The rewritten optimization problem has the right form to apply Prop.~\ref{th:kl-project-simplex} a second time. We obtain:
$$\textbf{x}^*_i = \frac{\textbf{q}_i}{\sum_{i'} \textbf{q}_{i'}}$$
By plugging this into $\textbf{Y}^*(\textbf{x})$, we obtain the solution to the overall optimization problem.

\subsection{Reduction of \cref{eq:min-cross-entropy} to \cref{eq:linear-program}}
\label{sec:argmax-and-kl}
In this section, we show how \cref{eq:min-cross-entropy} can be reduced to a problem of the entropy regularized version of \cref{eq:linear-program}. This is useful because it means we can use \cref{alg:kl-project} to solve \cref{eq:min-cross-entropy}.

First, we show that computing a KL projection is equivalent to solving an entropy-regularized linear program. Let $C$ be the feasible region of the linear constraints.
\begin{align*}
     & \argmax_{\textbf{x} \in C}  \textbf{s}^T \textbf{x} - \tau \sum_i \textbf{x}_i (\log \textbf{x}_i -1) \\
    &= \argmin_{\textbf{x} \in C} \tau \sum_i \textbf{x}_i (\log \textbf{x}_i -\frac{\textbf{s}_i}{\tau} - 1) \\
    &= \argmin_{\textbf{x} \in C} \kl(\textbf{x}\dbar\exp(\frac{\textbf{s}}{\tau})) \\
\end{align*}

Due to this, \cref{eq:min-cross-entropy} is equivalent to a linear program that has the same feasible region as \cref{eq:linear-program} except for the additional constraint $\textbf{U} \in \mathcal{Q}(\textbf{y},\textbf{z}')$. Note that $\textbf{U} \in \mathcal{Q}(\textbf{y},\textbf{z}')$ essentially rules out certain correspondences. Therefore we can approximately enforce $\textbf{U} \in \mathcal{Q}(\textbf{y},\textbf{z}')$ by masking $\textbf{U}^*(s)$ such that any forbidden correspondence receives a very low score.

\subsection{Derivation of loss function as ELBO}
\label{sec:loss-derivation}
We now show how the training procedure we use to train our permutation model can be derived from a form of evidence lower bound (ELBO).

Ideally, our permutation model would be a distribution $P_{\theta}(\textbf{R}|\textbf{x}, \textbf{z}')$ over permutation matrices $\textbf{R}$ and we would maximize the marginal likelihood, \ie marginalizing over all permutations:
\begin{align}
    P(\textbf{y}|\textbf{x}, \textbf{z}') = \sum_{\textbf{R} \in \Perm} P_{\theta}(\textbf{R}|\textbf{x}, \textbf{z}') P(\textbf{y}|\textbf{z}', \textbf{R}) \label{eq:marginal-log}
\end{align}
where $P(\textbf{y}|\textbf{z}', \textbf{R}) = \prod_j \sum_i R_{ij} \cdot \mathbb{1}(y_j = z_i)$ with $\mathbb{1}$ being the indicator function. $P(\textbf{y}|\textbf{z}', \textbf{R})$ returns 1 iff applying the permutation $\textbf{R}$ to $\textbf{z}'$ results in $\textbf{y}$. Unfortunately, computing \cref{eq:marginal-log} exactly is intractable in general due to the sum over permutation matrices. 
We instead use techniques from variational inference and consider the following evidence lower bound (ELBO):
\begin{align}
     \log P(\textbf{y}|\textbf{x}, \textbf{z}') \geq & \max_{Q} \expec{\textbf{R} \sim Q(\textbf{R}|\textbf{x},\textbf{z}',\textbf{y})} \log P(\textbf{y}|\textbf{z}', \textbf{R}) \nonumber \\
       & - \kl(Q(\textbf{R}|\textbf{x},\textbf{z}',\textbf{y})\dbar P_{\theta}(\textbf{R}|\textbf{x},\textbf{z}'))   \label{eq:tr-elbo}
\end{align}
where $Q(\textbf{R}|\textbf{x},\textbf{z}',\textbf{y})$ is an approximate variational posterior. We now relax the restriction that $P(\textbf{R}|\textbf{x},\textbf{z}')$ places non-zero mass only on permutation matrices and use the following definition of $P_{\theta}(\textbf{R}|\textbf{x},\textbf{z}')$:
\begin{align*}
        P_{\theta}(\textbf{R}_{ij} = 1|\textbf{x},\textbf{z}') = \textbf{U}^*(s)_{ij}
\end{align*}
where $\textbf{U}^*(s)$ is the solution to \cref{eq:linear-program} with added entropy regularization. 

It turns out, in our case, we can easily construct a variational posterior $Q$ that has zero reconstruction loss (the first term on the right side in \cref{eq:tr-elbo}): we can choose any $Q(\textbf{R}|\textbf{x},\textbf{z}',\textbf{y}) \in \mathcal{Q}(\textbf{y},\textbf{z}')$ where $\mathcal{Q}(\textbf{y},\textbf{z}')$ is the set of bistochastic matrices such that $Q(\textbf{R}|\textbf{x},\textbf{z}',\textbf{y})_{i,j} = 0$ iff  $\textbf{z}'_i \neq \textbf{y}_j$. To see that this gives zero reconstruction error, consider position $j$ in the output: The probability mass is distributed across precisely those positions $i$ in $\textbf{z}'$ where the right kind of token lives. In other words, any alignment with non-zero probability will reconstruct the output token at position $j$.

Therefore we can use the following lower bound to the log-likelihood:
\begin{align}
   & \log \: P(y|\textbf{x}, \textbf{z}') \geq \label{eq:min-kl} \\
   & - \min_{Q\in \mathcal{Q}(\textbf{y},\textbf{z})} \kl(Q(\textbf{R}|\textbf{x},\textbf{z}',\textbf{y}) \dbar P_{\theta}(\textbf{R}|\textbf{x},\textbf{z}')) \nonumber
\end{align}
During training, we need to compute the gradient of \cref{eq:min-kl}. 
By Danskin's theorem \citep{Danskin1967}, this is:
\begin{align}
    - \nabla_{\theta} \kl(Q^* \dbar  P_{\theta}(\textbf{R}|\textbf{x},\textbf{z}')) \label{eq:min-kl-grad}
\end{align}
where $Q^* \in \mathcal{Q}(\textbf{y},\textbf{z}')$ is the minimizer of \cref{eq:min-kl}. Note that $Q^*$ can equivalently be characterised as $\hat{\textbf{U}}$ (\cref{eq:min-cross-entropy}).

In practice, we also add $-\kl(\hat{\textbf{W}} | \textbf{W}^*(s))$ to our objective in \cref{eq:min-kl} to speed up convergence; this does not change the fact that we use a lower bound.

\section{Additional results and analysis}
\label{appendix:additional-results}

\begin{figure}[t]
    \centering
    \includegraphics[width=\linewidth]{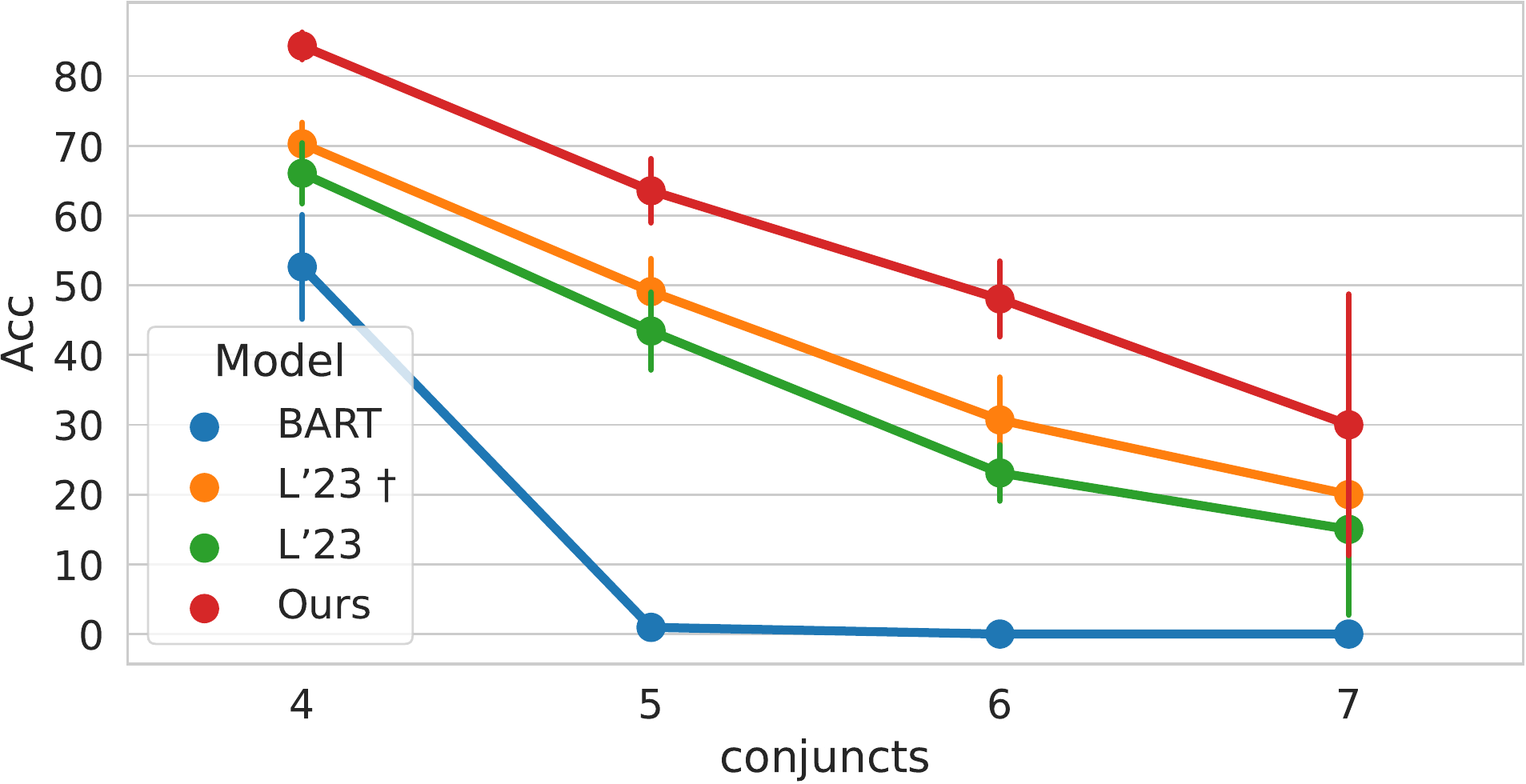}
    \caption{Accuracy on the document domain of Okapi by number of conjuncts in the gold logical form.}
    \label{fig:okapi-length}
\end{figure}

\paragraph{Okapi} In \cref{fig:okapi-length} we show the accuracy of our model on the document domain in comparison with previous work by number of conjuncts in the logical form.

\begin{table}[t]
    \centering
    \resizebox{0.95\linewidth}{!}{
\begin{tabular}{llll}
\toprule
& & \textbf{BART perm} & \textbf{Ours}  \\
\midrule
\multirow{3}{*}{COGS} & OS & 16\ci{8} & 33\ci{24} \\
& CP & 17\ci{3} & 82\ci{11} \\
& PP & 16\ci{5} & 91\ci{5} \\
\midrule
\multirow{2}{*}{ ATIS } & iid & 77.10\ci{1.61} & 76.65\ci{1.67}\\
& Length & 23.81\ci{9.45} & 41.39\ci{13.47} \\
\bottomrule
\end{tabular}}
    \caption{Performance of our permutation model in comparison to using BART for predicting the permutation. We ignore the order of conjuncts for evaluation.}
    \label{tab:bart-perm}
\end{table}

\paragraph{Permutation baseline} A simpler approach for predicting a permutation of the output $z'$ from the multiset tagging is to use a seq2seq model. In order to compare our approach to such a baseline, we concatenate the original input $x$ with a separator token and $z'$. We then feed this as input to a BART-base model which is trained to predict the output sequence $y$. At inference time, we use beam search and enforce the output to be a permutation of the input.
As detailed in \cref{tab:bart-perm}, this approach works well in-distribution and it also shows a small improvement over finetuning BART directly on the length split of ATIS. However, it does not perform as well as our approach. On COGS, our model outperforms the permutation baseline by an even bigger margin. Unseen variable symbols could be a challenge for BART on COGS which might explain part of the gap in performance.

This approach towards predicting permutations is similar to that of \citet{cazzaro2022translate} except that they do not constrain the beam search to permutations. 
We found that not constraining the output to be a permutation worked worse in the compositional generalization setups.

\section{Further model details}
\label{sec:model-details}

\subsection{Parametrization of permutation model}
We do not share parameters between the multiset tagging model and the permutation model.

Tokens that appear more than once in the same multiset have the same representation $\textbf{h}'_i$ in \cref{eq:hidden-rep-perm}. In order to distinguish them, we concatenate another embedding to $\textbf{h}'_i$: if the token $z'_i$ is the $k$-th instance of its type in its multiset, we concatenate an embedding for $k$ to $\textbf{h}'_i$. For example, in \cref{fig:intro}, $z'_5=\texttt{`}\texttt{x}_1\texttt{'}$ and it is the second instance of $\texttt{`}\texttt{x}_1\texttt{'}$ in its multiset, so we use the embedding for $2$.

We found it helpful to make the temperature $\tau$ of the scores for \cref{alg:kl-project} dependent on the number of elements in the permutation, setting $\tau = (\log n)^{-1}$, so that longer sequences have slightly sharper distributions.

Since the permutation model is designed to model exactly permutations, during training, $z$ and $y$ must have the same elements. This is not guaranteed because $z$ is the prediction of the multiset model which may not have perfect accuracy on the training data. For simplicity, we disregard instances where $z$ and $y$ do not have the same elements. In practice, this leads to a very small loss in training data for the permutation model.

\subsection{Lexicon mechanism}
\label{sec:lexicon}
The lexicon $L$ is a lookup table that deterministically maps an input token $x_i$ to an output token $L(x_i)$, and we modify the distribution for multiset tagging as follows:
\begin{align*}
  P'(\textbf{z}_{i,v}=k| \textbf{x}) = \begin{cases}
  P(\textbf{z}_{i,\mathscr{L}}=k| \textbf{x}_i) & \text{ if }v = L(\textbf{x}_i) \\
  P(\textbf{z}_{i,v}=k| \textbf{x})  & \text{ else}
  \end{cases}
\end{align*}
where $P(\textbf{z}_{i,v}=k| \textbf{x})$ is as defined in \cref{eq:multiset-dist} and $\mathscr{L}$ is a special lexicon symbol in the vocabulary. $P(\textbf{z}_{i,\mathscr{L}}| \textbf{x}_i)$ is a distribution over the multiplicity of $L(\textbf{x}_i)$, independent of the identity of $L(x_i)$. We use the `simple' lexicon induction method by \citet{akyurek-andreas-2021-lexicon}. Unless otherwise specified during learning, $L(\textbf{x}_i) = \textbf{x}_i$ like in a copy mechanism. 

\paragraph{Handling of variables in COGS}
For the COGS dataset, a model has to predict variable symbols. The variables are numbered (0-based) by the input token that introduced it (\eg in \cref{fig:intro}, slept, the third token, introduces a variable symbol $\texttt{x}_2$). In order to robustly predict variable symbols for sentences with unseen length, we use a similar mechanism as the lexicon look up table: we introduce another special symbol in the vocabulary, \texttt{Var}. If \texttt{Var} is predicted with a multiplicity of $k$ at $i$-th input token, it adds the token $\texttt{x}_{i-1}$ to its multiset $k$ times.

\subsection{Initialization of Multiset Tagging model}
\label{sec:init-multiset-model}
If there are $l$ alignments with a posterior probability of at least $\chi$ that an input token $i$ produces token $v$, we add the term $\lambda \log P(\textbf{z}_{i,v} \geq l\mid \textbf{x})$ to \cref{eq:multiset-objective}. $\lambda$ is the hyperparameter determining the strength. This additional loss is only used during the first $g$ epochs.

\section{Datasets and Preprocessing}
We show basic statistics about the data we use in \cref{tab:num-examples}. Except for the doubling task, all our datasets are in English. COGS uses a small fragment of English generated by a grammar, see \citet{kim-linzen-2020-cogs} for details.

\paragraph{Doubling task}
For the doubling task, we use an alphabet of size $|\Sigma| = 11$. To generate inputs with a specific range of lengths, we first draw a length from the range uniformly at random. The symbols in the input are also drawn uniformly at random and then concatenated into a sequence. Examples of lengths 5 - 10 are used as training, examples of length 11 are used as development data (e.g. for hyperparameter selection), and examples of length 11 - 20 are used as test data.

\subsection{Preprocessing}

\paragraph{COGS} Unlike \citet{zheng-lapata-2022-disentangled, qiu-etal-2022-improving, drozdov2022compositional} we do not apply structural preprocessing to the original COGS meaning representation and keep the variable symbols: all our preprocessing is local and aimed at reducing the length of the logical form (to keep runtimes low). We delete any token in $\{\texttt{",","\_","(",")","x",".",";","AND"}\}$ as these do not contribute to the semantics and can be reconstructed easily in post-processing. The tokens $\{$\texttt{"agent", "theme", "recipient", "ccomp", "xcomp", "nmod", "in", "on", "beside"}$\}$ are always preceded by a \texttt{"."} and we merge \texttt{"."} and any of those tokens into a single token.

Example:

\noindent
\texttt{* cookie ( x \_ 3 ) ; * table ( x \_ 6 ) ; lend . agent ( x \_ 1 , Dylan ) AND lend . theme ( x \_ 1 , x \_ 3 ) AND lend . recipient ( x \_ 1 , x \_ 9 ) AND cookie . nmod . beside ( x \_ 3 , x \_ 6 ) AND girl ( x \_ 9 )} 

\noindent
Becomes

\noindent
\texttt{* cookie 3 * table 6 lend .agent 1 Dylan lend .theme 1 3 lend .recipient 1 9 cookie .nmod .beside 3 6 girl 9}

\paragraph{ATIS} We follow the pre-procressing by \citet{fertility-paper} and use the variable-free FunQL representation as annotated by \citet{guo-etal-2020-benchmarking-meaning}. We use spacy \texttt{3.0.5} (model \texttt{en\_core\_web\_sm}) to tokenize the input.

\paragraph{Okapi} Again, we follow the preprocessing of \citet{fertility-paper}. We use spacy \texttt{3.0.5} (model \texttt{en\_core\_web\_sm}) to tokenize both the input utterances and the output logical forms.

\begin{table}[t]
    \centering
\resizebox{\linewidth}{!}{
\begin{tabular}{llrrr}
\toprule
\textbf{Dataset} & \textbf{Split/Version} & \textbf{Train} & \textbf{Dev} & \textbf{Test} \\
\midrule
Doubling & & 4,000 & 500 & 1,000 \\
\midrule
COGS & & 24,155 & 3,000 & 21,000 \\
\midrule
\multirow{2}{*}{ ATIS } & iid & 4,465 & 497 & 448 \\
& length & 4,017 & 942 & 331 \\
\midrule
\multirow{3}{*}{ Okapi} & Calendar & 1,145 & 200 & 1061 \\
& Document & 2,328 & 412 & 514 \\
& Email & 2,343 & 200 & 991 \\
\midrule
\end{tabular}
}
    \caption{Number of examples per dataset/split.}
    \label{tab:num-examples}
\end{table}

\section{Details on evaluation metrics}
We provide code for all evaluation metrics in our repository.

\paragraph{Doubling} We use exact match accuracy on the string.

\paragraph{COGS} For COGS we use exact match accuracy on the sequence in one evaluation setup. The other evaluation setup disregards the order of conjuncts: we first remove the `preamble' (which contains all the definite descriptions) from the conjunctions. We count a prediction as correct if the set of definite descriptions in the preamble matches the set of definite descriptions in the gold logical form \textit{and} the set of clauses in the prediction match the set of clauses in the gold logical form.

\paragraph{ATIS}
We allow for different order of conjuncts between system output and gold parse in computing accuracy. We do this by sorting conjuncts before comparing two trees node by node. This is the same evaluation metric as used by \citet{fertility-paper}.

\paragraph{Okapi}
We follow \citet{okapi, fertility-paper} and disregard the order of the parameters for computing accuracy. We use a case-insensitive string comparison.

\section{Hyperparameters}
We use the same hyperparameters for all splits of a dataset. For our model, we only tune the hyperparameters of the multiset tagging model; the permutation model is fixed, and we use the same configuration for all tasks where we use RoBERTa. For model ablations where we use an LSTM instead of RoBERTa, we use the same hyperparameters for Okapi and ATIS, and a smaller model for the doubling task. These configurations were determined by hand without tuning.
For BART, we use the same hyperparameter as \citet{fertility-paper}.

We follow the random hyperparameter search procedure of \citet{fertility-paper} for the multiset tagging models and the LSTM/transformer we train from scratch: we sample 20 configurations and evaluate them on the development set. We run the two best-performing configurations again with a different random seed and pick the one with the highest accuracy (comparing the union of the predicted multisets with the gold multiset). We then train and evaluate our model with entirely different random seeds.

The chosen hyperparameters along with the search space are provided in the github repository.

\section{Number of parameters, computing infrastructure and runtime}
We show the number of parameters in the models we train in \cref{tab:num-params}.

All experiments were run on GeForce GTX 1080 Ti or GeForce GTX 2080 Ti with 12GB RAM and Intel Xeon Silver or Xeon E5 CPUs.

The runtime of one run contains the time for training, evaluation on the devset after each epoch and running the model on the test set. We show runtimes of the model we train in \cref{tab:runtimes}. Since we evaluate on 5 random seeds (10 for COGS due to high variance of results), our experiments overall took around 64 hours of compute time on our computing infrastructure.

\begin{table}[]
    \centering
    \resizebox{\linewidth}{!}{
\begin{tabular}{llll}
\toprule
\textbf{Dataset} & \textbf{Model} & \textbf{First stage} & \textbf{Second stage} \\
\midrule
\multirow{3}{*}{ Doubling } & LSTM & 5 & - \\
& Transformer & 3 & - \\
& Ours & 3 & 14$^*$ \\
\midrule
COGS & Ours & 20 & 80 \\
\midrule
\multirow{2}{*}{ ATIS } & Ours & 30 & 50 \\
& Ours/LSTM & 30 & 110 \\
\midrule
\multirow{2}{*}{ Okapi/Calendar } & Ours & 9 & 35 \\
& Ours/LSTM & 5 & 30 \\
\midrule
\multirow{2}{*}{ Okapi/Email } & Ours & 12 & 40 \\
& Ours/LSTM & 9 & 40 \\
\midrule
\multirow{2}{*}{ Okapi/Document } & Ours & 12 & 70 \\
& Ours/LSTM & 10 & 55 \\
\bottomrule
\end{tabular}}
    \caption{Average runtime for train/evaluate on dev and test in \textbf{minutes}. For the doubling task, we note that our model has converged usually after $\frac{1}{4}$ of the time in the table.}
    \label{tab:runtimes}
\end{table}

\begin{table}[t]
	\centering
	\resizebox{\linewidth}{!}{
\begin{tabular}{llll}
	\toprule
	\textbf{Dataset} & \textbf{Model} & \textbf{First stage} & \textbf{Second stage} \\
	\midrule
	\multirow{3}{*}{ Doubling } & LSTM & & 2.462 \\
	& Transformer & & 10.424 \\
	& Ours & 0.273 & 0.463 \\
	\midrule
	COGS & Ours & 125.091 & 127.16 \\
	\midrule
	\multirow{2}{*}{ ATIS } & Ours & 125.506 & 127.119 \\
	& Ours/LSTM & 3.345 & 1.816 \\
	\midrule
	\multirow{2}{*}{ Okapi/Calendar } & Ours & 124.927 & 127.03 \\
	& Ours/LSTM & 1.493 & 1.876 \\
	\bottomrule
\end{tabular}
	}
	\caption{Number of parameters in millions in the models we train. This includes the $124.646$m params of RoBERTa when we finetune it.}
	\label{tab:num-params}
\end{table}

\end{document}